\documentclass[preprint,12pt]{elsarticle}

\usepackage{amssymb}
\usepackage{amsmath}

\usepackage[colorlinks,urlcolor=blue,linkcolor=blue,citecolor=blue]{hyperref}

\usepackage{color,array}

\usepackage{graphicx}
\usepackage{amsmath,amsfonts}
\usepackage{url}

\usepackage[ruled,vlined,linesnumbered]{algorithm2e}

\usepackage{array}
\usepackage[caption=false,font=normalsize,labelfont=sf,textfont=sf]{subfig}
\usepackage{textcomp}
\usepackage{stfloats}
\usepackage{url}
\usepackage{verbatim}
\usepackage{graphicx}
\usepackage{cite}
\usepackage{cleveref}

\usepackage{amsmath}
\usepackage{amsfonts}
\usepackage{graphicx}
\usepackage{booktabs}
\usepackage{hyperref}
\usepackage{color}

\newcommand{\lstm}{\texttt{LSTM}}
\newcommand{\mlp}{\texttt{MLP}}
\newcommand{\dnn}{\texttt{DNN}}
\newcommand{\rnn}{\texttt{RNN}}
\newcommand{\trans}{\texttt{Transformer}}

\journal{Energy and AI}

\begin{document}

\begin{frontmatter}



\title{Climate Aware Deep Neural Networks (CADNN) for Wind Power Simulation} 





\author[UFZ]{Ali Forootani*}
\ead{ali.forootani@ufz.de/aliforootani@ieee.org}
\author[UFZ]{Danial Esmaeili~Aliabadi}
\ead{danial.esmaeili@ufz.de}
\author[UFZ,DBFZ,UNI]{Daniela Thr\"an}
\ead{daniela.thraen@ufz.de}
\address[UFZ]{Helmholtz Centre for Environmental Research - UFZ, Permoserstraße 15, 04318 Leipzig, Germany}
\address[DBFZ]{DBFZ Deutsches Biomasseforschungszentrum gGmbH, Torgauer Strasse 116, 04347 Leipzig, Germany}
\address[UNI]{University Leipzig, Institute for Infrastructure and Resources Management, Grimmaische Str. 12, 04109 Leipzig, Germany}
\cortext[cor1]{Corresponding author}


\begin{abstract}
Wind power forecasting plays a critical role in modern energy systems, facilitating the integration of renewable energy sources into the power grid. Accurate prediction of wind energy output is essential for managing the inherent intermittency of wind power, optimizing energy dispatch, and ensuring grid stability. This paper proposes the use of Deep Neural Network (\dnn)-based predictive models that leverage climate datasets, including wind speed, atmospheric pressure, temperature, and other meteorological variables, to improve the accuracy of wind power simulations.

In particular, we focus on the Coupled Model Intercomparison Project (CMIP) datasets, which provide climate projections, as inputs for training the \dnn~models. These models aim to capture the complex nonlinear relationships between the CMIP-based climate data and actual wind power generation at wind farms located in Germany. Our study compares various \dnn~architectures, specifically Multilayer Perceptron (\mlp), Long Short-Term Memory (\lstm) networks, and \trans-enhanced \lstm~models, to identify the best configuration among these architectures for climate-aware wind power simulation.

The implementation of this framework involves the development of a Python package (\texttt{CADNN}) designed to support multiple tasks, including statistical analysis of the climate data, data visualization, preprocessing, \dnn~training, and performance evaluation. We demonstrate that the \dnn~models, when integrated with climate data, significantly enhance forecasting accuracy. This climate-aware approach offers a deeper understanding of the time-dependent climate patterns that influence wind power generation, providing more accurate predictions and making it adaptable to other geographical regions.  
\end{abstract}


\begin{highlights}
\item Climate data-driven DNNs boost wind power forecasting accuracy
\item LSTM and Transformer models excel in climate-aware wind simulations.
\end{highlights}

\begin{keyword}
Coupled Model Intercomparison Project (CMIP), Deep Neural Network (\dnn), Wind Power, Climate Dataset, Long Short Term Memory (\lstm)-\dnn, .



\end{keyword}

\end{frontmatter}



\section{Introduction}
{T}{he} role of renewable energy (RE) in reducing greenhouse gas (GHG) emissions is crucial, which is essential for mitigating the impact of climate change. Europe’s 2030 Climate and Energy Strategy aims to reduce GHG emissions by $40\%$ compared to $1990$ levels and achieve a $27\%$ share of renewable energy in total electricity consumption by $2030$ \citet{oberthur2019hard}. Wind power is one of the leading renewable energy source in terms of installed capacity and growth, contributing $15\%$ of the European Union’s electricity consumption in $2019$ \citet{asiaban2021wind,zhang2024high}. However, wind power is highly sensitive to climate change, as future changes in wind patterns can significantly impact electricity production due to the cubic relationship between wind speed and energy potential \citet{carvalho2017potential}. Wind farms typically have lifetimes of $20–30$ years, making it essential to estimate future changes in wind resources under climate change scenarios. Factors, such as mean wind speeds and inter-annual variability, can affect the reliability and profitability of wind energy\citet{carvalho2012ocean,pryor2010climate}.

Global Climate Models (GCMs), particularly from the Coupled Model Intercomparison Project (CMIP), provide crucial data for assessing climate change impacts \citet{ahmadalipour2017multi}. CMIP is a collaborative framework designed to compare and improve climate models by coordinating experiments across multiple research institutions \citet{eyring2016overview}. The project has become a cornerstone in climate science, offering standardized simulations that help evaluate models' performance and project future climate scenarios. CMIP involves numerous coupled models that simulate interactions between different Earth system components, such as the atmosphere, oceans, land surface, and sea ice, enabling the study of feedbacks and interactions driving climate change \citet{lovato2022cmip6}. Each CMIP phase builds on the previous one: CMIP3 supported the IPCC Fourth Assessment Report (AR4) in 2007, CMIP5 contributed to the Fifth Assessment Report (AR5) in 2013, and CMIP6, the latest phase, was used for the IPCC Sixth Assessment Report (AR6) with a more comprehensive scope, including new model components, higher resolution, and complex interactions between human activities and climate \citet{eyring2016overview}.

Studies using CMIP5 data generally show an increase in wind resources in northern Europe and a decrease in southern Europe, with increased seasonality but inconclusive evidence for inter-annual variability \citet{reyers2016future}. Recent reviews and studies have analyzed climate change impacts on wind energy using CMIP and CORDEX data, indicating a decrease in offshore wind energy resources for Europe, except in northern Europe and parts of the Iberian Peninsula \citet{decastro2019overview, santos2018accuracy, costoya2020using}. Similar studies for the US, China, and Africa reveal region-specific changes, with projected decreases for many coastal areas and some increases in inland or specific offshore regions \citet{chen2020impacts, costoya2020suitability, costoya2021climate}. For instance, the Guinea coast of West Africa is expected to see a rise in wind energy resources, while the Sahel region may experience a decrease \citet{akinsanola2021projected}. To make matters more complicated, the arrangement of wind farms can reduce the capacity factor of its downwind turbines and neighboring downwind farms by $20\%$ \citet{akhtar2021accelerating}. 

The methodologies used in these studies involve multi-model ensembles (\texttt{MME}s) validated with reanalysis data, such as ERA5, to ensure accuracy \citet{soares2019climate}. Despite advances in resolution and agreement between models, challenges remain, particularly regarding the spatial resolution of future climate projections and their impacts on accurately predicting RE production using less computational resources. Therefore, it is essential for revisiting previous research using CMIP6 data, which incorporates more realistic future scenarios, and notes that improvements in GCMs are expected to enhance wind resource predictions for Europe and other regions \citet{carvalho2021wind}.


\section{literature review}
The growing significance of wind power prediction in recent years is largely due to its applications at a commercial scale, offering a clean alternative to conventional energy sources. Early work in \citet{brown1984time} introduced time-series forecasting for wind power. Since then, various researchers have proposed regression-based methods for more efficient wind power prediction. For example, \citet{liu2012short} presented a short-term wind power forecasting system that incorporates wavelet transforms, turbine attributes, and Support Vector Machines (\texttt{SVM}). The authors introduced a technique that combines data mining and \texttt{K-means} clustering to analyze wind power data, removing redundant data before forecasting using an Artificial Neural Network (\texttt{ANN}).

\citet{yuan2015short} proposed a hybrid approach, where the parameters of a Least Square Support Vector Machine (\texttt{LSSVM}) are optimized using Gravitational Search Algorithms (\texttt{GSA}). \citet{lee2013short} offered another short-term forecasting approach, training 52 \texttt{ANN}s and five Gaussian sub-models in parallel using historical data. These models generate power forecasts for the same time period, and a final prediction is made by aggregating their outputs. \citet{pinson2009skill} suggested a method based on ensemble predictions, while \citet{de2011error} trained models using real wind farm data from three turbines, utilizing multiple models, including Auto regression moving average (\texttt{ARMA}), \texttt{ANNs}, and Adaptive Neuro-Fuzzy Inference Systems.

\citet{bhaskar2012awnn} developed an efficient wind power prediction system comprising two phases: the first phase applies wavelet decomposition to wind series, and the second phase uses a neural network to transform wind speed into power. Similarly, \citet{abedinia2015short} presented a system that uses \texttt{ANN}, where the optimal number of hidden neurons is determined through a heuristic method, given that \texttt{ANN} performance often depends on hidden layer size. \citet{mabel2008analysis} also applied \texttt{ANN} to predict wind power based on three years of data from seven wind farms. \citet{chitsaz2015wind} proposed a system that combines clonal search algorithms and wavelet neural networks during training. \citet{grassi2010wind} introduced a unique approach using a two-layer \texttt{ANN} with three neurons in each hidden layer and a single output neuron to predict wind farm energy. The first layer uses a hyperbolic tangent activation function, while the second hidden layer uses a sigmoid function.

The 2012 Global Energy Forecasting Competition showcased several methods for reliable wind power prediction \citet{hong2014global}, in which the utilized datasets were also described. \citet{lei2009review} provided another review of techniques related to wind power forecasting. Researchers have reported various hybrid approaches for wind power forecasting \citet{bao2014short, costa2008review, foley2012current}, and a comprehensive review of hybrid techniques is presented by \citet{tascikaraoglu2014review} proposed a method based on action-dependent heuristic dynamic programming, using a radial basis function neural network (\texttt{RBFNN}) to predict maximum wind power. Another hybrid method, introduced by \citet{peng2013hybrid}, combines statistical and physical techniques for wind power forecasting. \citet{bao2014short} used Empirical Mode Decomposition (\texttt{EMD}) to break wind power into high- and low-frequency components, with Relevance Vector Machines (\texttt{RVM}) applied to forecast these frequencies. \citet{pinson2009probabilistic} employed statistical methods to enhance wind power predictions by converting prediction errors into multivariate Gaussian random variables. \citet{nielsen2006using} extended recent wind power prediction systems by proposing quantile regression for probabilistic forecasting.

Ensemble-based methods have also been explored in the literature for wind speed prediction.\citet{salcedo2009accurate} utilized a bank of \texttt{ANN}s, demonstrating that their approach outperformed single neural network models for wind speed prediction. \citet{troncoso2015local} employed various regression trees to predict wind speeds across multiple wind farms and compared the results with those from \texttt{ANN}s and Support Vector Regression (\texttt{SVR}). \citet{ortiz2011short} proposed an ensemble technique using a bank of \texttt{SVR}s to forecast wind speed, outperforming \mlp s. \citet{zhang2013short} presented a hybrid model for wind speed forecasting using data from northwest China. To address the limitations of \texttt{ANN}s in forecasting, \citet{ma2017generalized} developed a model based on dynamic fuzzy \texttt{ANN}s. Despite the success of statistical and hybrid approaches, each has its own limitations \citet{jung2014current}.

Although these models are accurate enough to address operational needs based on past data, the underlying climate system is changing, and hence, these models may not be fit anymore for future needs \citet{esmaeili4960744climate}. To this end, in this manuscript, we are preparing the ground for the next generation of \texttt{ANN}-based models that can utilize the results of climate models. 

Moreover, Previous efforts to estimate wind power generation using physics-based models were hindered by the significant variability in climate scenarios and the limitations of current climate models in providing scenario-specific data with high spatial and temporal resolution \citet{lehneis2022power}. To address these challenges, we have shifted from physical simulation models to employing \dnn~architectures, leveraging the processed CMIP climate dataset to enhance the accuracy of wind power predictions.

\subsection{Contributions}  
This article introduces a refined methodology for integrating CMIP6 climate data into wind energy forecasting over Germany. The proposed approach enhances key aspects such as data preparation, spatial interpolation, temporal alignment, and scaling, enabling effective localized forecasts at wind turbine sites. By tailoring CMIP6 data for machine learning applications, we bridge the gap between large-scale climate models and localized renewable energy prediction. Key Methodological Advancements are as follows:
\begin{itemize}  
    \item Data Preparation:  
    \begin{itemize}  
        \item Temporal resampling and conversion to a standardized time format ensure compatibility with time-series forecasting models, particularly deep learning architectures like \lstm~networks that rely on consistent temporal structures for accurate predictions.  
        \item Feature normalization is applied to harmonize input scales, promoting balanced contributions during model training and improving predictive performance.  
    \end{itemize}  

    \item Spatial Interpolation:
    An interpolation algorithm accurately maps rotated pole datasets to geographic coordinates, ensuring precise spatial alignment with wind conditions across Germany—a critical step for localized forecasting.  

    \item Model Evaluation:
    This study systematically evaluates multiple deep learning architectures, including Multi-Layer Perceptrons (\mlp s), \lstm~networks, and \trans-enhanced \lstm s:  
    \begin{itemize}  
        \item \mlp s effectively capture static relationships but struggle with the temporal and non-linear dependencies inherent in climate data.  
        \item \trans-enhanced \lstm s, despite their complexity, did not yield significant performance improvements, indicating a mismatch with the dataset’s characteristics.  
        \item \lstm~networks outperform others due to their robust ability to model sequential and long-term dependencies, making them the preferred architecture for climate-driven renewable energy forecasting.  
    \end{itemize}  
\end{itemize}  
 
To facilitate reproducibility, we developed a dedicated Python package, \texttt{CADNN}, leveraging \texttt{PyTorch} for deep learning implementation. This package includes modules for data preparation, model training, and evaluation, optimized for large climate datasets with temporal and spatial components. By sharing this package, we provide the research community with a robust toolset for implementing \dnn-based wind power simulation, fostering accessible and efficient workflows.

To the best of our knowledge, this is the first study to integrate CMIP6 climate data into localized wind power simulation at this scale. Our findings highlight \lstm~networks as the most effective architecture for climate-based renewable energy prediction, offering valuable insights and establishing best practices for preparing and utilizing climate data in machine learning applications.


This paper is organized as follows: In \cref{background} we briefly provide a background about \dnn~models that we make use in this article. We analyze the meteorological data in \cref{analy_data} in CMIP data files over Germany in the locations of wind farms. In \cref{dnn_climate} we investigate the performance of three \dnn~architectures in mapping nonlinear relationship of climate dataset and the wind power. The conclusion is given in \cref{conclude}.


\section{Mathematical Background of \dnn s} \label{background}

A \dnn~is a type of \texttt{ANN}~with multiple layers of neurons. The primary components of a \dnn~are neurons, layers, weights, and activation functions \citet{forootani2011fault}.

\paragraph{Mathematical Representation}  
Let \( \mathbf{x} \) represent the input vector, and \( \mathbf{W}_l, \mathbf{b}_l \) represent the weight matrix and bias vector of the \( l \)-th layer, respectively. The output of the \( l \)-th layer is computed as:
\[
\mathbf{z}_l = \mathbf{W}_l \mathbf{a}_{l-1} + \mathbf{b}_l,
\]
where \( \mathbf{a}_{l-1} \) is the activation from the previous layer. The activation function \( \sigma \) is applied to produce the output of the layer:
\[
\mathbf{a}_l = \sigma(\mathbf{z}_l),
\]
Common choices for activation functions include: (i) Sigmoid: \( \sigma(x) = \frac{1}{1 + e^{-x}} \); (ii) ReLU: \( \sigma(x) = \max(0, x) \); (iii) Tanh: \( \sigma(x) = \tanh(x) \).

The final output layer of the network computes the predicted value:
\[
\mathbf{y}_{\text{pred}} = \mathbf{W}_L \mathbf{a}_{L-1} + \mathbf{b}_L,
\]

\paragraph{Training}  
\dnn s are typically trained using back-propagation and an optimization algorithm (usually stochastic gradient descent, SGD). The loss function \( L(\mathbf{y}_{\text{pred}}, \mathbf{y}_{\text{true}}) \) measures the difference between the predicted and true outputs. The goal of training is to minimize the loss by adjusting the weights and biases through gradient descent:
\[
\mathbf{W}_l \leftarrow \mathbf{W}_l - \eta \frac{\partial L}{\partial \mathbf{W}_l},
\]
where \( \eta \) is the learning rate.

\subsection{\lstm~Networks}

\lstm~networks are a type of Recurrent Neural Network (\rnn) designed to address the vanishing gradient problem in traditional \rnn s, particularly for sequences with long-range dependencies \citet{huang2022well}.

\paragraph{\lstm~Architecture}  
An \lstm~unit consists of several gates: input gate, forget gate, and output gate. The cell state \( \mathbf{C}_t \) and hidden state \( \mathbf{h}_t \) are updated at each time step. The gates in \lstm~are computed as follows:
\\
- Forget gate:
\[
\mathbf{f}_t = \sigma(\mathbf{W}_f [\mathbf{h}_{t-1}, \mathbf{x}_t] + \mathbf{b}_f),
\]
- Input gate:
\[
\mathbf{i}_t = \sigma(\mathbf{W}_i [\mathbf{h}_{t-1}, \mathbf{x}_t] + \mathbf{b}_i),
\]
\[
\mathbf{\tilde{C}}_t = \tanh(\mathbf{W}_C [\mathbf{h}_{t-1}, \mathbf{x}_t] + \mathbf{b}_C),
\]
- Output gate:
\[
\mathbf{o}_t = \sigma(\mathbf{W}_o [\mathbf{h}_{t-1}, \mathbf{x}_t] + \mathbf{b}_o),
\]
where $\sigma$ is the activation function \footnote{ The candidate cell state (\(\mathbf{\tilde{C}}_t\)) and the final cell state modulation (\(\tanh(\mathbf{C}_t)\)) use \(\tanh\) because the cell state stores the actual information (or ``content'') of the memory}. The cell state is updated as:
\[
\mathbf{C}_t = \mathbf{f}_t \odot \mathbf{C}_{t-1} + \mathbf{i}_t \odot \mathbf{\tilde{C}}_t,
\]
where \( \odot \) denotes element-wise multiplication.

The hidden state is updated as:
\[
\mathbf{h}_t = \mathbf{o}_t \odot \tanh(\mathbf{C}_t).
\]

\paragraph{Training}  
Similar to \dnn s, \lstm s are trained via backpropagation through time (BPTT), where gradients are computed for each time step and used to update the weights using an optimization algorithm like SGD or Adam.

\subsection{ \trans-based \dnn s}

\trans~networks are primarily used for sequence-to-sequence tasks such as machine translation, and text summarization \citet{vaswani2017attention}. \trans~utilize self-attention mechanisms to model dependencies between elements in a sequence \citet{forootani2024bio}.

\paragraph{\trans~Architecture}  
The \trans~model consists of an encoder and a decoder, both composed of stacked layers of self-attention and feed-forward neural networks.

\paragraph{Self-Attention}  
For each input sequence, the self-attention mechanism computes a scaled dot-product attention for each query \( \mathbf{q}_i \), computed as:
\[
\text{Attention}(\mathbf{q}_i) = \text{softmax}\left(\frac{\mathbf{q}_i \mathbf{k}_j^T}{\sqrt{d_k}}\right) \mathbf{v}_j,
\]
where \( \mathbf{q}_i \) is the query vector for the \(i\)-th element, \( \mathbf{k}_j \) is the key vector for the \(j\)-th element, \( \mathbf{v}_j \) is the value vector for the \(j\)-th element, and \( d_k \) is the dimension of the key vector.

The self-attention mechanism captures dependencies by weighting the importance of each element in the sequence with respect to others.

\paragraph{Multi-Head Attention}  
The model applies multiple attention mechanisms in parallel (multi-head attention) to capture different aspects of dependencies.

\paragraph{Position Encoding}  
Since \trans s do not have an inherent notion of sequence order, position encodings are added to the input to provide information about the position of tokens in the sequence. The position encoding \( \mathbf{p}_t \) for position \( t \) is typically computed as:
\[
\mathbf{p}_t = [\sin(\frac{t}{10000^{\frac{2i}{d}}}), \cos(\frac{t}{10000^{\frac{2i+1}{d}}})] \quad \text{for} \quad i = 1, 2, \dots, d,
\]
where \( d \) is the dimension of the input.

\paragraph{Feed-Forward Network}  
After the attention mechanism, the output is passed through a feed-forward neural network:
\[
\mathbf{y}_i = \text{ReLU}(\mathbf{W}_2 (\text{ReLU}(\mathbf{W}_1 \mathbf{x}_i + \mathbf{b}_1)) + \mathbf{b}_2),
\]

\paragraph{Training}  
\trans~are trained using maximum likelihood estimation and the cross-entropy loss function, with optimization algorithms like Adam.

Summary of Key Differences for different \dnn~structures are as follows:
\begin{itemize}
    \item \textbf{\mlp s}: General-purpose neural networks that model complex relationships via multiple layers of neurons and nonlinear activation functions.
    \item \textbf{\lstm s}: A type of \rnn~designed to handle long-range dependencies in sequential data using gating mechanisms to manage memory over time.
    \item \textbf{\trans s}: Sequence models based on self-attention mechanisms that capture long-range dependencies more effectively than \rnn~s and \lstm s, especially in tasks like machine translation.
\end{itemize}


\section{Analysis of Meteorological Data Over Germany}\label{analy_data}
Effective data processing techniques are crucial for managing and analyzing large-scale datasets in scientific and industrial domains, particularly in renewable energy and climate studies. Techniques such as data cleaning, validation, interpolation, and machine learning integration are essential for extracting actionable insights from atmospheric and geospatial data, including wind speed, pressure, and temperature. Publicly accessible datasets like those from CMIP6 require spatial interpolation to estimate values at specific locations, using methods like the \emph{Regular Grid Interpolator} for geographic grids \citet{jung2014current, weiser1988note}, and temporal interpolation to handle missing time-series data. Climate models often provide results with a lower temporal resolution (typically a 3-hour resolution\footnote{\url{https://cds.climate.copernicus.eu/datasets}}), while wind power data is recorded hourly or even quarter-hourly (e.g., Open Power System Data\footnote{\url{https://open-power-system-data.org/}}). To align these datasets, a resampling and aggregation approach is necessary, transforming wind power and pressure data into standard intervals, such as 3-hour periods, to match the climate data.

In this context, accurately preparing wind speed and surface pressure datasets for any region\footnote{Germany is selected as a case study in this manuscript.} involves extracting and processing CMIP6 data. This preparation includes spatial interpolation, scaling, and temporal resampling to ensure the data is suitable for wind power simulation models and other atmospheric analyses. To capture fine spatial variations, latitude and longitude coordinates are defined at a specific resolution, and geographical ranges relevant to the selected region (i.e., Germany in this study) are filtered. This approach streamlines the CMIP data, yielding a focused dataset that supports robust analysis of wind speed and pressure in the region, facilitating effective decision-making in renewable energy forecasting and climate studies.

In climate data analysis, especially when using rotated pole datasets such as CMIP, transforming coordinates and spatially interpolating data onto a target region (e.g., Germany) are critical for ensuring that the extracted variables accurately represent the geographical area of interest. The following two steps—coordinate transformation and spatial interpolation—are implemented to achieve this precision.

\paragraph{Coordinate Transformation}
The CMIP datasets utilize a rotated pole coordinate system where geographic coordinates are defined relative to an artificial ``rotated'' pole. This coordinate system is commonly used to reduce grid distortion in regional climate modeling \citet{grose2023cmip6}. To align with standard geographic coordinates (latitude and longitude), a coordinate transformation is required. Using the Algorithm \cref{alg:rotate_coordinates}, we convert the standard latitude and longitude of target points within Germany to this rotated coordinate system. This ensures that each point of interest in Germany is accurately aligned with the dataset’s coordinate system before data extraction, maintaining the spatial integrity of climate variables such as wind speed.

\paragraph{Spatial Interpolation and Efficient Spatial Querying with \texttt{KD-Tree} on the climate dataset}
After transforming coordinates, the next challenge is to spatially match these target points to the dataset's grid points, which may not align perfectly. The \texttt{cKDTree} is a binary space-partitioning data structure used to organize points \(\mathbf{V} = \{\mathbf{v}_1, \mathbf{v}_2, \dots, \mathbf{v}_n\} \in \mathbb{R}^k\), where each node splits the space at a median point along one of the \(k\) dimensions. At depth \(d\), the space is split along dimension \(d \mod k\) using the median value, \(\mathbf{v}_\text{split} = \text{median}(v_1^{(d \mod k)}, \dots, v_n^{(d \mod k)})\), dividing points into two subsets. For nearest-neighbor search, given a query point \(\mathbf{q} \in \mathbb{R}^k\), it efficiently finds the closest point \(\mathbf{v}_\text{nearest} = \arg \min_{\mathbf{v}_i \in \mathbf{V}} \|\mathbf{q} - \mathbf{v}_i\|_2\). The \texttt{KD-tree} accelerates queries like nearest neighbors or radius searches by pruning irrelevant parts of the space, reducing search complexity to \(O(\log n)\) on average \footnote{Here, the ``cKDTree'' algorithm from ``scipy.spatial'' is applied to efficiently locate the nearest dataset grid points for each transformed coordinate \citet{virtanen2020scipy}.}.

Constructing a \texttt{KD-Tree} for the grid enables rapid spatial querying, significantly reducing the computational complexity compared to exhaustive search methods. This function then performs a nearest-neighbor search, returning the closest available grid point to each transformed target coordinate. This approach minimizes interpolation errors and ensures that the extracted data points accurately represent spatial locations within the target region.

This combination of coordinate transformation and \texttt{KD-Tree} spatial querying provides an accurate and computationally efficient methodology for extracting climate variables at the desired spatial resolution. It enhances the fidelity of regional climate analysis by aligning extracted data points with geographic coordinates in a physically meaningful way.

In conclusion the \texttt{KD-Tree} algorithm is used to efficiently map the extracted wind speeds or pressure surface from the rotated grid of the CMIP6 dataset to a regular latitude-longitude grid over Germany, ensuring spatial data consistency during the extraction process. However we need another interpolation algorithm to further refine this data by interpolating the wind speeds or pressure surface onto specific, irregularly spaced target locations (e.g., measurement sites), addressing the mismatch between the extracted grid and the desired points of analysis.

\paragraph{Spatial Interpolation at the location of wind turbines}

To refine the spatial representation of both wind speed and pressure data, interpolation techniques are applied. Wind speeds and pressure levels at the locations of wind farms are interpolated onto the desired target locations (such as wind turbine sites or measurement stations). In particular, we employ a linear interpolation method (\emph{Regular Grid Interpolator}), which performs multivariate interpolation on a regular grid in \( n \)-dimensional space. Suppose the function \( f: \mathbb{R}^n \to \mathbb{R} \) is known on a set of regularly spaced grid points \( \{x_1^i, x_2^j, \dots, x_n^k\} \), where each \( x_i \in X_i \) forms a regular grid for each dimension. The objective is to estimate \( f(\mathbf{x}) \), where \( \mathbf{x} = (x_1, x_2, \dots, x_n) \), which is not necessarily a grid point. 

To interpolate at \( \mathbf{x} \), let \( (x_1^i, x_1^{i+1}), \dots, (x_n^k, x_n^{k+1}) \) be the grid intervals that contain \( x_1, \dots, x_n \). For linear interpolation in each dimension, the value of \( f(\mathbf{x}) \) is obtained as a weighted combination of the function values at the corners of the grid cell surrounding \( \mathbf{x} \). Denote the neighboring grid points as \( f(x_1^i, x_2^j, \dots, x_n^k) \), and the interpolated value is given by:

\[
f(\mathbf{x}) \approx \sum_{i_1=0}^1 \sum_{i_2=0}^1 \dots \sum_{i_n=0}^1 w_{i_1 i_2 \dots i_n} f(x_1^{i_1}, x_2^{i_2}, \dots, x_n^{i_n}),
\]
where \( w_{i_1 i_2 \dots i_n} \) are the weights based on the relative distances between \( \mathbf{x} \) and the surrounding grid points. For example, for linear interpolation in 1-D, the interpolation formula between two grid points \( x_i \) and \( x_{i+1} \) is:

\[
f(x) \approx \frac{x_{i+1} - x}{x_{i+1} - x_i} f(x_i) + \frac{x - x_i}{x_{i+1} - x_i} f(x_{i+1}).
\]

In cases where interpolated values can not be computed due to a lack of surrounding data, the nearest available data points were used as a fallback, ensuring that the final interpolated dataset contained no gaps. Interpolation allows for higher spatial granularity of both wind speed and pressure data, enhancing the model’s ability to predict localized weather conditions. Algorithm \cref{alg:meteorological_interpolation} summarizes the interpolation method that is explained earlier.

\paragraph{Temporal Resampling}

Wind power production usually is measured hourly\footnote{\url{https://open-power-system-data.org/}}, however both wind speed and surface pressure in CMIP data are available in $3$-hour interval, therefore the power is resampled. This approach retains essential temporal variability while reducing computational requirements for subsequent analysis. To ensure consistency across the dataset, the datetime information is standardized, facilitating time-based calculations in machine learning workflows. This transformation enables the data to be effectively utilized in time-series forecasting models, ensuring that temporal patterns are accurately captured. These steps of resampling and time standardization are crucial for preparing the data for machine learning applications, such as \lstm~networks, which depend on a regular time structure to enhance forecasting accuracy in wind power prediction.

\paragraph{Data Scaling and Normalization}

Data scaling is an essential step in preparing datasets for machine learning models, which can be sensitive to the range and distribution of input values. Here, wind speed and pressure data were scaled using a \texttt{Min-Max} scaling technique to map values between $-1$ and $1$. This approach ensures that each feature contributes equally to the learning process, avoiding biases from variables with larger numerical ranges.

Normalizing variables like wind speed, pressure, and time ensures that machine learning models can identify patterns effectively, without being affected by differences in scale across features. This scaling process enhances the performance of machine learning algorithms, such as \dnn s, which typically perform better when input features are standardized.

\begin{algorithm}
    \caption{Rotate Coordinates}
    \label{alg:rotate_coordinates}
    \KwIn{Latitude $\phi$, Longitude $\lambda$, Pole Latitude $\phi_p$, Pole Longitude $\lambda_p$ (in degrees)}
    \KwOut{Rotated Latitude $\phi_{rot}$, Rotated Longitude $\lambda_{rot}$ (in degrees)}
    
    \textbf{Convert coordinates to radians:} \\
    \[
    \phi \gets \frac{\pi}{180} \cdot \phi, \quad \lambda \gets \frac{\pi}{180} \cdot \lambda, \quad \phi_p \gets \frac{\pi}{180} \cdot \phi_p, \quad \lambda_p \gets \frac{\pi}{180} \cdot \lambda_p
    \]
    
    \textbf{Calculate longitude difference:} \\
    \[
    \Delta \lambda \gets \lambda - \lambda_p
    \]

    \textbf{Compute rotated latitude:} \\
    \[
    \phi_{rot} \gets \arcsin\left(\sin(\phi) \cdot \sin(\phi_p) + \cos(\phi) \cdot \cos(\phi_p) \cdot \cos(\Delta \lambda)\right)
    \]

    \textbf{Compute rotated longitude:} \\
    Define:

    \begin{align*}
       y &= \cos(\phi) \cdot \sin(\Delta \lambda),\\
       \quad x &= \sin(\phi) \cdot \cos(\phi_p) - \cos(\phi) \cdot \sin(\phi_p) \cdot \cos(\Delta \lambda)
    \end{align*}

    Then calculate:
    \[
    \lambda_{rot} \gets \lambda_p + \text{atan2}(y, x)
    \]
    where:
    \[
    \text{atan2}(y, x) = 
    \begin{cases} 
        \arctan\left(\frac{y}{x}\right) + \pi & \text{if } x < 0 \text{ and } y \geq 0 \\ 
        \arctan\left(\frac{y}{x}\right) + 2\pi & \text{if } x < 0 \text{ and } y < 0 \\ 
        \arctan\left(\frac{y}{x}\right) & \text{if } x > 0 \\ 
        \frac{\pi}{2} & \text{if } x = 0 \text{ and } y > 0 \\ 
        -\frac{\pi}{2} & \text{if } x = 0 \text{ and } y < 0 \\ 
        0 & \text{if } x = 0 \text{ and } y = 0 
    \end{cases}
    \]

    \textbf{Convert rotated coordinates back to degrees:} \\
    \[
    \phi_{rot} \gets \frac{180}{\pi} \cdot \phi_{rot}, \quad \lambda_{rot} \gets \frac{180}{\pi} \cdot \lambda_{rot}
    \]
    
    \KwRet{$(\phi_{rot}, \lambda_{rot})$}
\end{algorithm}


\begin{algorithm}
\caption{Interpolation of Meteorological Data}
\label{alg:meteorological_interpolation}
\KwIn{Meteorological data \(\mathbf{X} \in \mathbb{R}^{T \times M \times N}\) (e.g., wind speeds or pressures), grid latitudes \(\mathbf{L}_{\text{lat}} \in \mathbb{R}^M\), grid longitudes \(\mathbf{L}_{\text{lon}} \in \mathbb{R}^N\), target points \(\mathbf{P} \in \mathbb{R}^{P \times 2}\)}
\KwOut{Interpolated meteorological values at target points \(\mathbf{X}_{\text{interp}} \in \mathbb{R}^{T \times P}\)}

\textbf{Sort grid latitudes and longitudes:} \\
\[
\mathbf{L}_{\text{lat}} \gets \text{sort}(\mathbf{L}_{\text{lat}}), \quad \mathbf{L}_{\text{lon}} \gets \text{sort}(\mathbf{L}_{\text{lon}})
\]

\textbf{Initialize matrix for interpolated values:} \\
\[
\mathbf{X}_{\text{interp}} \gets \text{zeros}(T, P)
\]

\For{\textbf{each} target point \(\mathbf{p} \in \mathbf{P}\) \textbf{and each time step} \(t\)}{
    \textbf{Reshape} meteorological data \(\mathbf{X}_t\) into a 2D grid.\\
    
    \textbf{Linear interpolation:} \\
    Define:
    \[
    \mathbf{X}_{\text{linear}}(\mathbf{p}) = \sum_{i=1}^{2} \sum_{j=1}^{2} w_{ij} \cdot \mathbf{X}(\mathbf{L}_{\text{lat}}[x_i], \mathbf{L}_{\text{lon}}[y_j])
    \]
    where \(w_{ij}\) are interpolation weights, and \((x_i, y_j)\) are indices of the grid points around \(\mathbf{p}\).
    
    \If{\text{linear interpolation fails to find values in} \(\mathbf{X}_{\text{linear}}\)}{
        \textbf{Nearest-neighbor interpolation:} \\
        Define:
        \[
        \mathbf{X}_{\text{nearest}}(\mathbf{p}) = \mathbf{X}(\mathbf{L}_{\text{lat}}[x_{\text{nearest}}], \mathbf{L}_{\text{lon}}[y_{\text{nearest}}])
        \]
        where \((x_{\text{nearest}}, y_{\text{nearest}})\) are indices of the closest grid points to \(\mathbf{p}\).
    }
    
    \textbf{Store} interpolated values for the current time step in \(\mathbf{X}_{\text{interp}}\).
}
\KwRet{\(\mathbf{X}_{\text{interp}}\)}
\end{algorithm}


\subsection{Dataset Visualization}

\begin{figure*}[!t]
\centering
\includegraphics[width=0.6\linewidth]{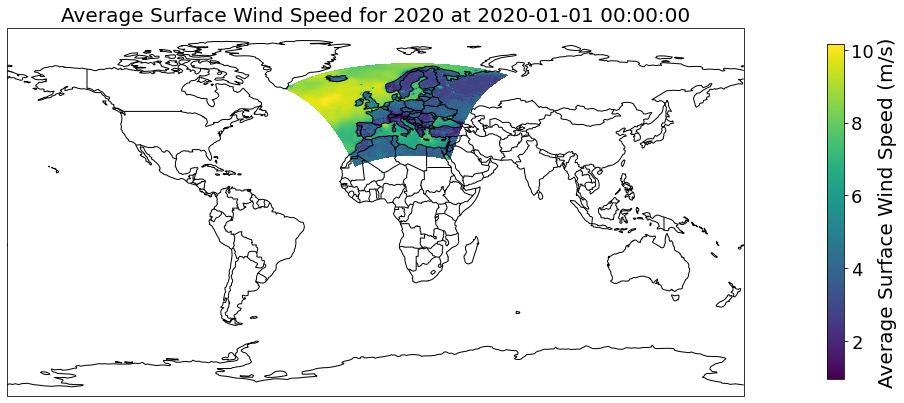}%

\caption{Average surface wind speed for Europe on 2020-01-01 at 00:00:00 from CMIP dataset.}
\label{EU_ave_wind_speed}
\end{figure*}

To facilitate interpretation, various visualization tools are used. In Figure \ref{EU_ave_wind_speed}, we present the average surface wind speed across European countries based on the CMIP dataset. We apply \cref{alg:rotate_coordinates} to transform the coordinates, followed by \cref{alg:meteorological_interpolation} to interpolate meteorological data specifically for the locations of wind farms in Germany. Notably, the locations of these wind farms are sourced from the EE-Monitor project,\footnote{EE Monitor provides information on the expansion of renewable energies in Germany from an environmental perspective, \url{https://web.app.ufz.de/ee-monitor/}.} which provides environmental insights into renewable energy expansion in Germany.

Wind speed and surface pressure data, each with a shape of $(2928\times28574)$, have been extracted from the CMIP dataset. Here, \(2928\) represents the number of time steps at 3-hour intervals across one year, and \(28574\) represents the various grid points. These grid points correspond to a latitude and longitude grid with shapes \((157)\) and \((182)\), respectively. 

To focus on specific target locations, wind speed and pressure data are interpolated, resulting in arrays with shapes of $(2928 \times 232)$ for both variables. The target points, which represent the locations of wind farms in Germany, are provided as specific coordinates of interest with a shape of $(232 \times 2)$. This data structure highlights the challenge of high dimensionality within climate datasets.

\begin{figure}[!htb]
    \centering
    \includegraphics[scale=0.23]{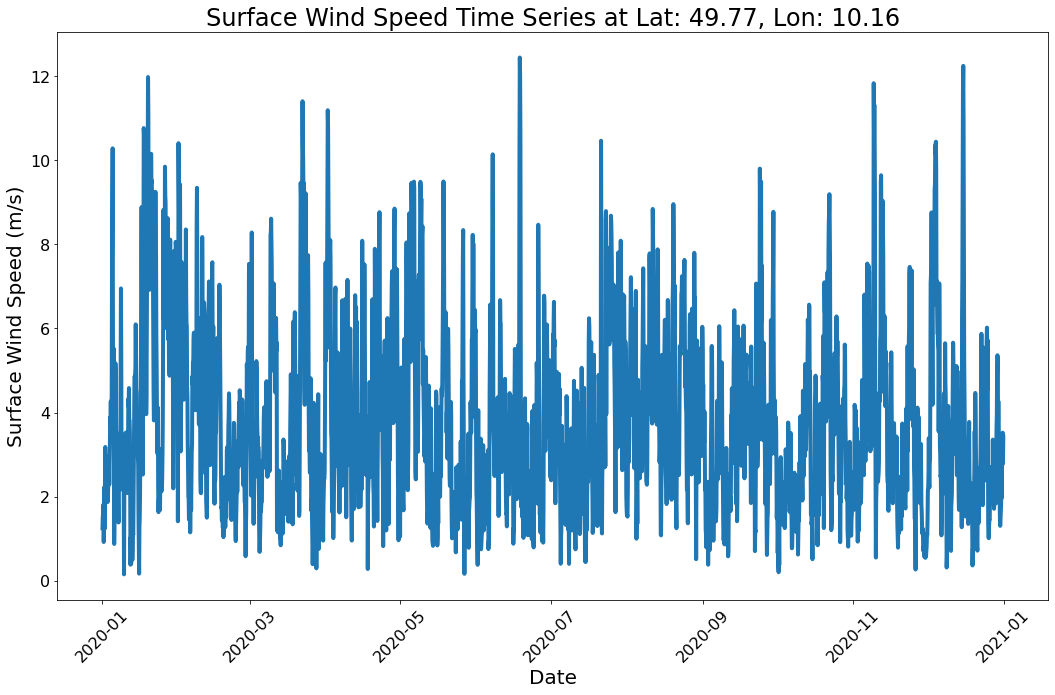}
    \caption{Surface Wind Speed for the entire year 2020 at coordinates: \( \text{Lat} \, 49.77, \, \text{Lon} \, 10.16 \) correspond to the city of W\"{u}rzburg in Germany.}
    \label{Wur_wind_time_series}
\end{figure}

In Figure \ref{Wur_wind_time_series}, the surface wind speed is shown at a location with coordinates \( \text{Lat} \, 49.77, \, \text{Lon} \, 10.16 \) correspond to the city of W\"{u}rzburg in Germany for the year 2020. Also, Figure \ref{interpolated_wind_speed} visualizes spatial variations in mean wind speeds across a grid of target locations, providing a comprehensive representation of wind dynamics in Germany. Color intensity indicates average wind speed values, where areas with higher mean speeds are highlighted to aid in assessing wind power potential and regional variability over the observed period. 

\begin{figure*}[!t]
\centering
\subfloat[]{\includegraphics[scale=0.25]{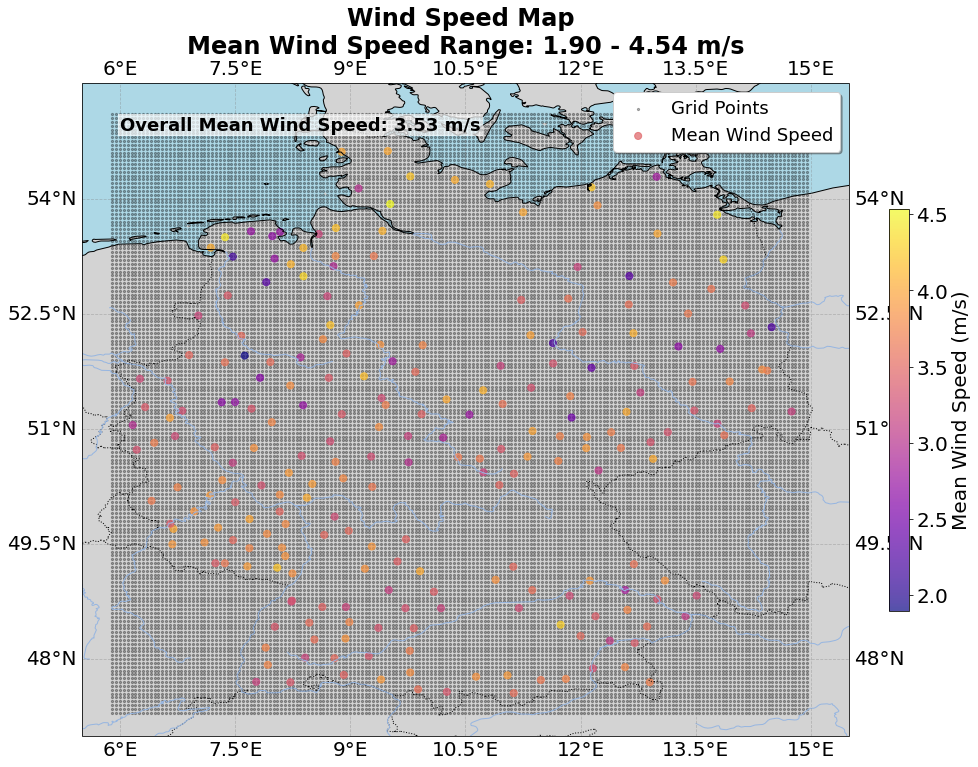}%
\label{interpolated_wind_speed}}
\hfil
\subfloat[]{\includegraphics[scale=0.25]{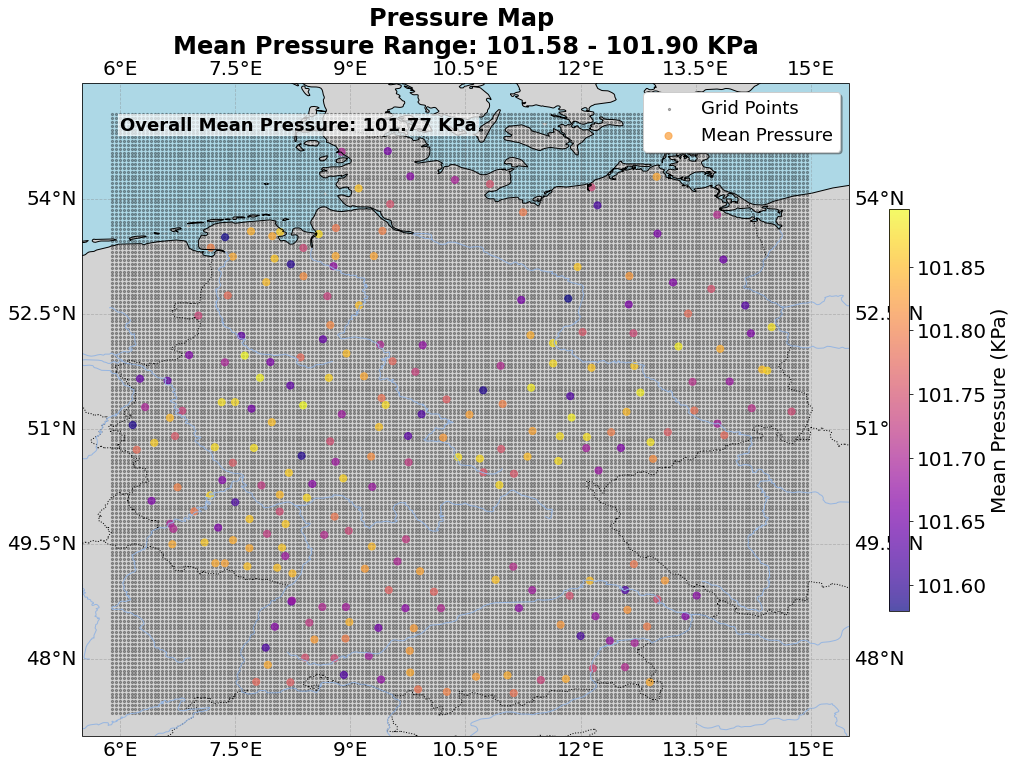}%
\label{interpolated_pressure}}
\caption{(a) Interpolation of wind speed over Germany computed by \cref{alg:rotate_coordinates} and \cref{alg:meteorological_interpolation} at the locations of wind farms. (b) Interpolation of pressure surface over Germany computed by \cref{alg:rotate_coordinates} and \cref{alg:meteorological_interpolation} at the locations of wind farms.}
\label{fig_sim}
\end{figure*}

The map in Figure \ref{interpolated_pressure} illustrates the spatial distribution of mean atmospheric pressure across Germany, where color intensity reflects pressure levels in kilopascals (KPa). This visualization enables an analysis of regional pressure variations and trends, which are essential for understanding atmospheric conditions relevant to weather forecasting and climate studies.


\begin{figure}
	\centering
	\includegraphics[scale=0.22]{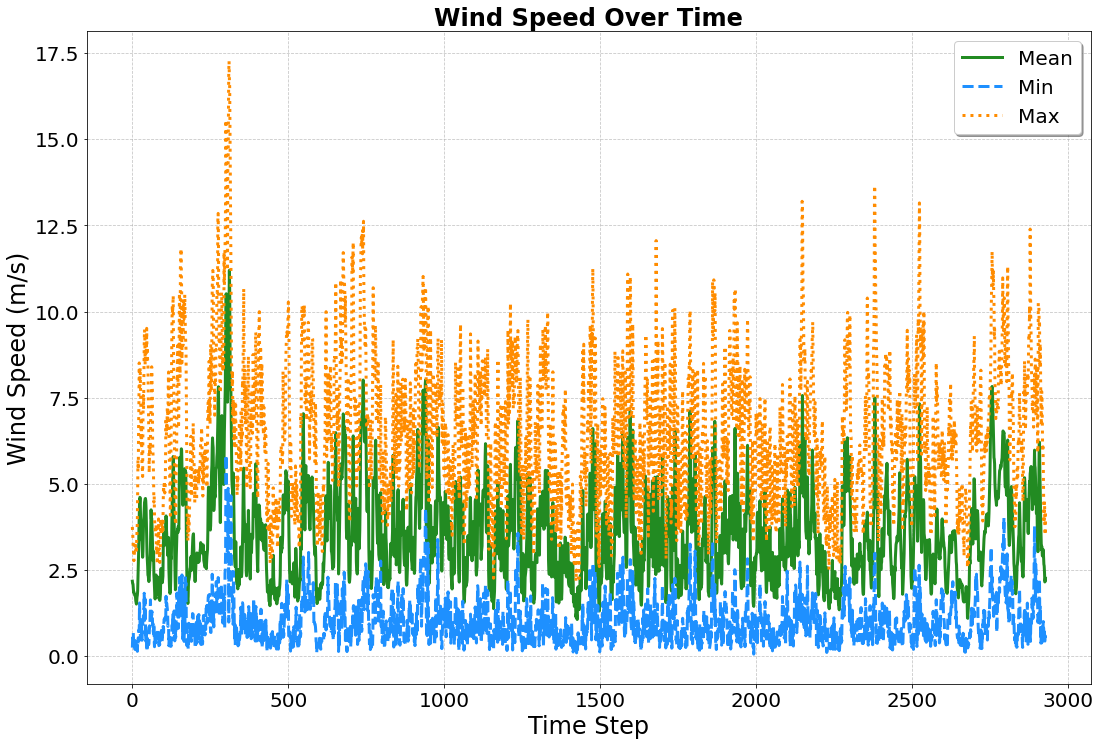} 
 \caption{Wind speed with the resolution of $3$ hours at locations of operational wind power plants in 2020.}
	\label{interpolated_wind_speed}
\end{figure}

\begin{figure}
	\centering
	\includegraphics[scale=0.22]{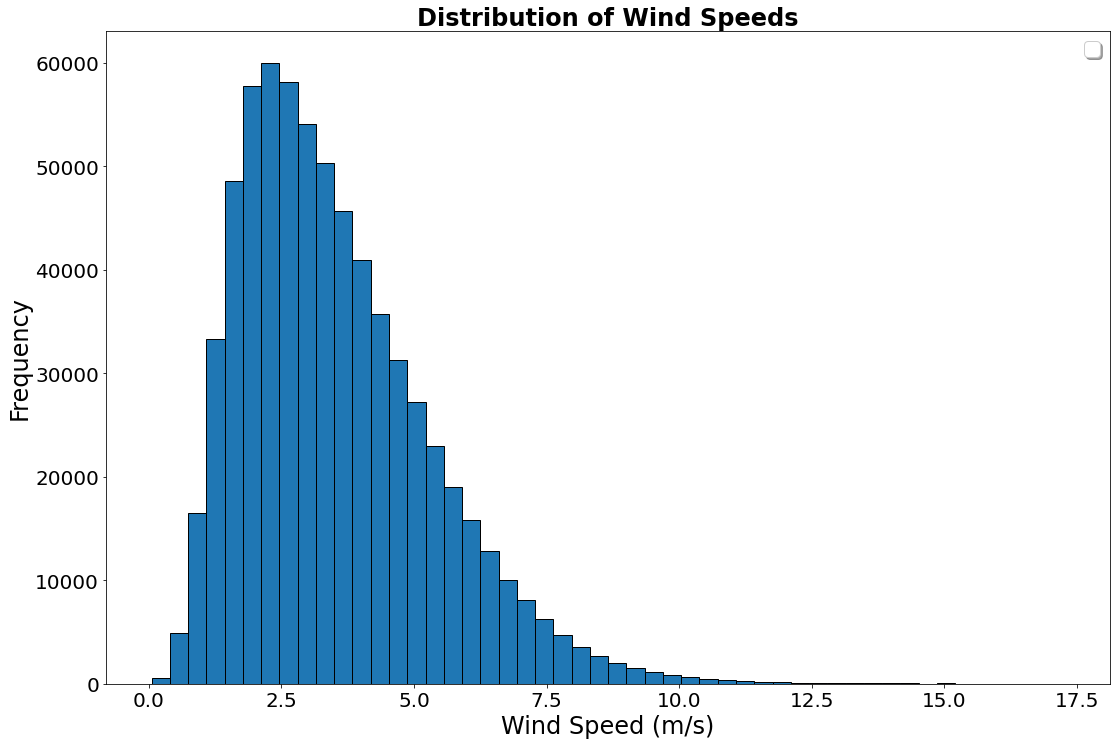}\caption{Distribution of wind speeds across the dataset for the year 2020.}
	\label{wind_speed_histogram}
\end{figure}

The histogram in \cref{wind_speed_histogram} illustrates the distribution of wind speeds across the dataset, highlighting the frequency of various wind speed intervals. This distribution analysis provides insight into prevalent wind conditions, essential for assessing wind energy potential and variability.

\begin{figure}
	\centering
	\includegraphics[scale=0.22]{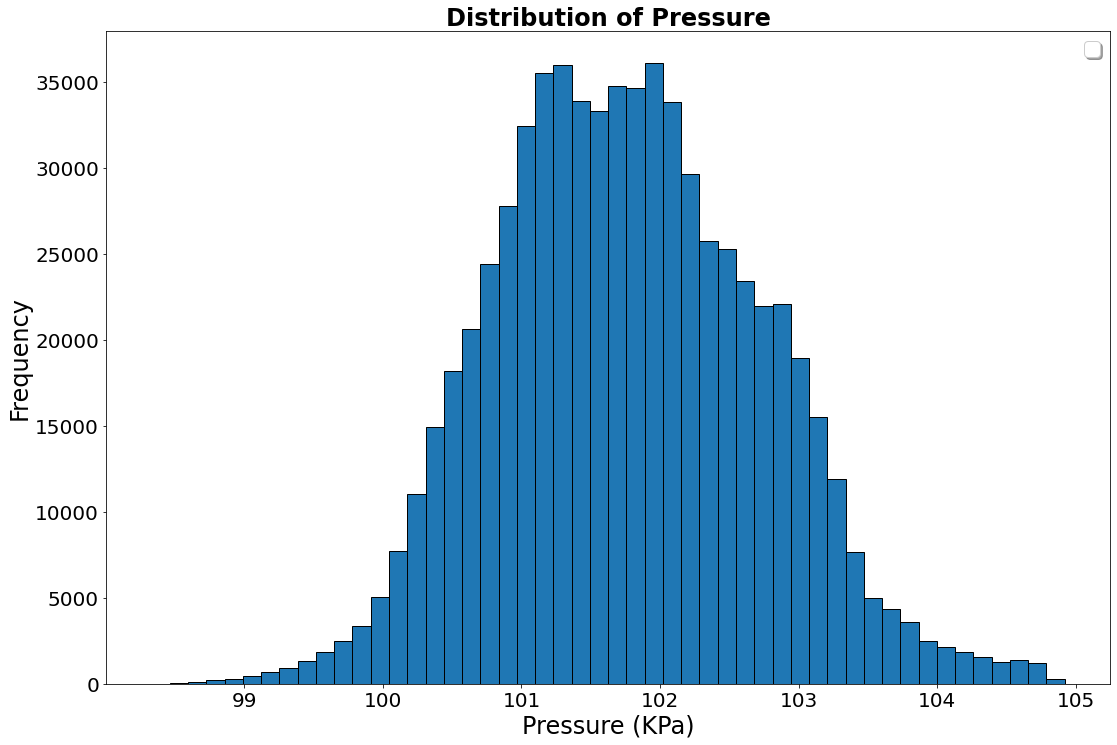}\caption{Distribution of pressure surface across the dataset for the year 2020.}
	\label{pressure_histogram}
\end{figure}

The histogram in \cref{pressure_histogram} displays the distribution of atmospheric pressure values across the dataset, converted to kilopascals (\texttt{KPa}). This frequency analysis of pressure variations aids in understanding atmospheric patterns, which is crucial for weather prediction and climate-related studies.


\begin{figure}
	\centering
	\includegraphics[scale=0.23]{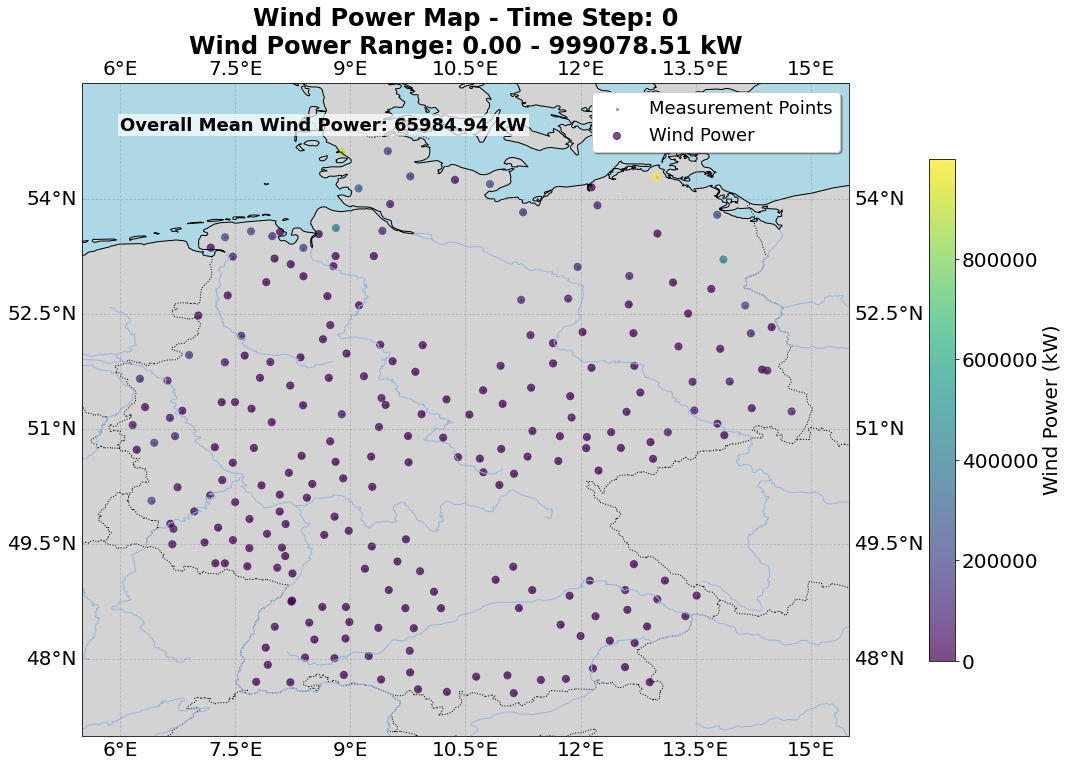}\caption{Wind power generation power in 2020 at time step 00:00:00, with resolution of $3$ hours.}
	\label{real_wind_power}
\end{figure}


\section{\dnn s for Wind Power Production Forecast Enhanced with Climate Data} \label{dnn_climate}

Previous attempts to generate a high-quality estimation of wind power generation based on climate models using physics-aware models did not yield acceptable results due to the inherent variabilities of climate scenarios and the inability of current climate models to offer scenario-based data with high spatial and temporal sensitivity \citet{lehneis2022power}. Thus, instead of using physical simulation models, we employ \dnn~architectures on the processed CMIP climate dataset to improve prediction accuracy in wind power simulating. Specifically, we compare three types of \dnn~models—\mlp s, \lstm~networks, and \trans-enhanced \lstm~networks—to evaluate their performance and suitability for this task.

Advancing wind power simulating with climate data requires machine learning models capable of capturing complex, non-linear patterns and temporal dependencies. \dnn s are well-suited for this purpose, as they can adapt to a wide range of input structures and feature relationships. \mlp s are particularly effective for capturing non-linear input-output mappings in high-dimensional data, making them useful for static feature extraction from climate variables such as wind speed and surface pressure. However, as \mlp s are fully connected networks without memory structures, they are less effective at modeling sequential dependencies over time. This limitation can affect their accuracy in time-sensitive forecasting applications.

In contrast, \lstm~networks, a type of recurrent neural network (\rnn), are designed to capture time dependencies through their unique gating mechanisms and memory cells. This structure enables them to retain information over extended periods, allowing \lstm~networks to model sequential relationships and capture historical trends that influence future outputs. While \lstm s require more computational resources and training time, they often yield better performance on tasks involving time series data, such as wind power simulating based on climate variables.

To further enhance the temporal modeling capabilities, we include a \trans-enhanced \lstm~model. The \trans~component is integrated to focus attention on significant past events in the time series, selectively weighting them based on relevance to current predictions. This hybrid approach combines the sequential memory strength of \lstm s with the attention mechanisms of \trans s, allowing for dynamic weighting of historical patterns over varying time scales. This architecture is expected to improve forecast accuracy, especially in cases where specific climate conditions have a disproportionate impact on wind power production.

\begin{figure*}[htp]
\centering
{\includegraphics[width=5.5in]{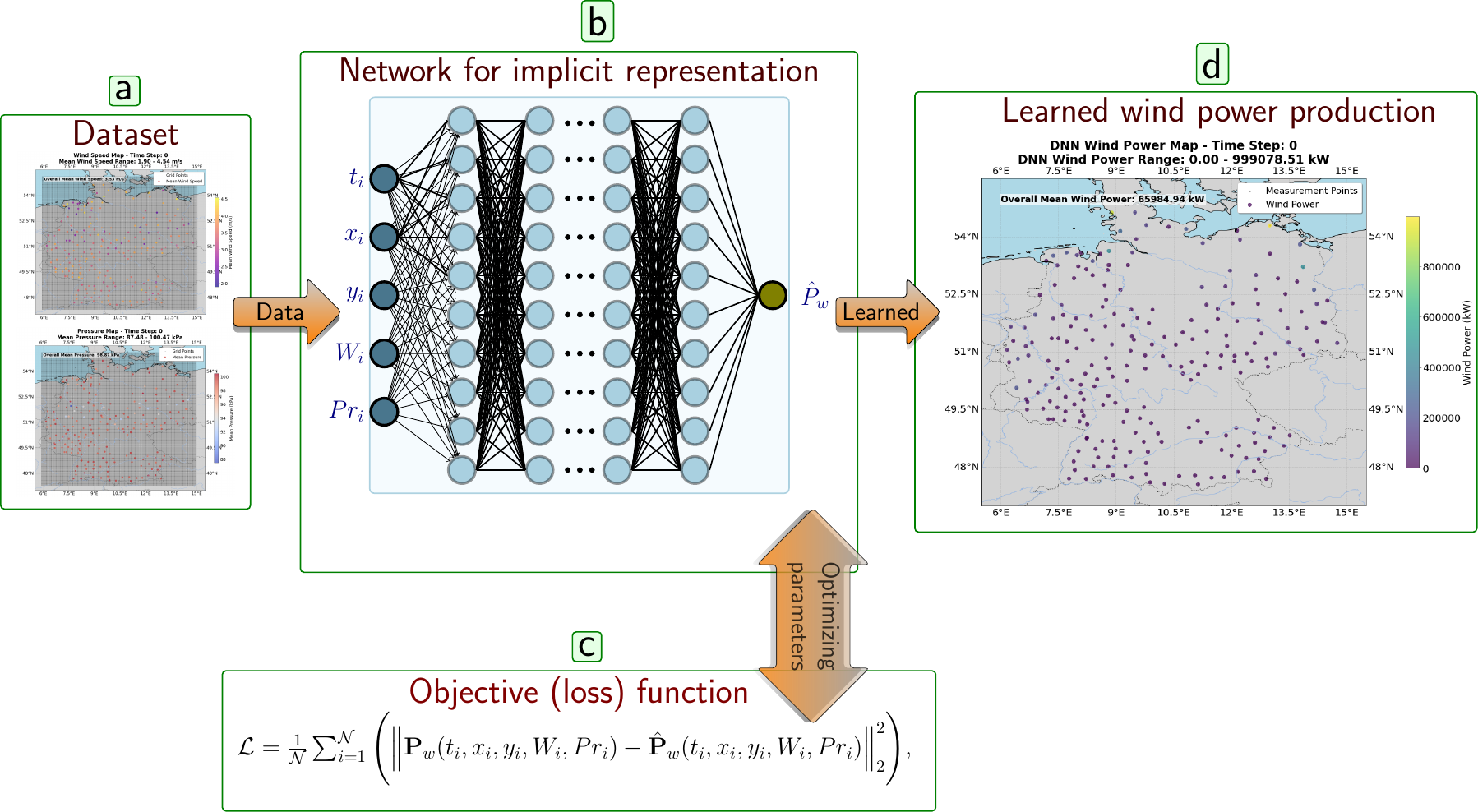}%
}
\caption{Deep learning architecture for wind power simulation. (a) climate dataset including wind speed and pressure surface; (b) \mlp~based \dnn~architecture takes 5 inputs, time $t$, $x$ and $y$ coordinates, wind speed $W$ and pressure surface $Pr$; (c) training of the \dnn~based on the mean square loss function; (d) learned wind power production.}
\label{MLP_DNN_wind_train}
\end{figure*}

In Figure \ref{MLP_DNN_wind_train}, we illustrate the \mlp-based \dnn~architecture for wind power prediction, showcasing the structure of the model, the input climate features, and the training process. By analyzing the performance of \mlp, \lstm, and \trans-enhanced \lstm~models, we can assess the trade-offs between model complexity, computational cost, and predictive accuracy. This comparison provides insights into the optimal architecture for wind power simulating using processed CMIP climate datasets, helping to determine the most effective approach for leveraging both static and temporal features in climate-driven prediction tasks.

In contrast, \lstm-\dnn s are specifically designed to manage sequential data, making them highly effective for tasks where time series analysis is crucial. The unique architecture of \lstm s, with memory cells and gating mechanisms, allows them to retain information over long periods and capture dependencies that are spread out across time. This ability makes \lstm s particularly suited for scenarios where past climate conditions impact current or future wind power outputs. While \lstm-\dnn s tend to have higher computational costs and require more time to train compared to \mlp s, they often offer superior performance in forecasting tasks that involve sequential or time-dependent data.

By comparing these architectures, we can better understand which model offers improved predictive capabilities when applied to processed CMIP climate datasets for wind power simulating, considering the trade-off between model complexity, training time, and accuracy.

Unlike \mlp-\dnn s, \lstm-\dnn~requires more sophisticated dataset preparation, therefore in the next subsection we provide its associated details.

\subsection{Data Preparation Process for \lstm-Based Time Series wind power simulation}


Preparing data for \lstm-based deep learning models is essential for capturing temporal dependencies in time series forecasting. This process involves four key stages: partitioning the dataset, generating sequences, reshaping the data, and initializing data loaders. Each stage ensures that the model receives well-structured inputs, enabling it to learn from historical data and generalize effectively to new, unseen data.

The first stage is to divide the dataset into training and testing subsets based on a predefined ratio of test data. This split is carried out chronologically to maintain the sequential nature of the data and avoid potential data leakage between the training and testing phases. This step is critical for ensuring that the model is evaluated on data that it has not seen during training, which is necessary for obtaining an unbiased assessment of its predictive performance.

Next, fixed-length sequences are created from the training and testing datasets. These sequences consist of consecutive data points over a specified length, providing the \lstm~model with historical context that is crucial for making accurate predictions of future time steps. A rolling window approach is employed, extracting sequences of data followed by the corresponding target value for prediction.

Once the sequences are generated, they must be reshaped into the appropriate format for \lstm~input, typically \texttt{Batch size}, \texttt{Sequence Length}, \texttt{number of features}. This reshaping step ensures that the data is compatible with the \lstm~architecture, enabling efficient computation and facilitating parallel processing during model training.

Finally, data loaders are employed to manage the input data during training. By enabling mini-batch processing, data loaders improve computational efficiency, particularly when training on \texttt{GPU}s, and help accelerate the convergence of the model. Additionally, shuffling within batches promotes better generalization, reducing the risk of overfitting and improving the model’s ability to make reliable predictions on unseen data.

\cref{alg:lstm_data_preparation} summarizes this data preparation process, providing a structured approach for training \lstm-based models in the context of wind power simulation.


\begin{algorithm}
\caption{Generalized Data Preparation for \lstm~with Random Sampling}
\label{alg:lstm_data_preparation}
\KwIn{\\ \texttt{Data} - Complete dataset with features and target values;\\
\texttt{Coordinates} - Optional spatial or feature coordinates;\\ 
\texttt{Sequence Length} - Desired length of input sequences for \lstm;\\ \texttt{Test Data Size} - Fraction of data to be used for testing;\\
\texttt{Batch Size} - Number of samples per batch in data loader;\\
\texttt{Device} - \texttt{cuda} if available, else \texttt{cpu}}
\KwOut{\\ 
Training sequences, Testing sequences, and their corresponding data loaders}

\textbf{Set device based on \texttt{GPU} availability:} \\
\[
\texttt{device} \gets \texttt{cuda} \text{ if available, else } \texttt{cpu}
\]

\textbf{Step 1: Train-Test Split} \\
\texttt{Partition} \texttt{Data} into \texttt{Train} and \texttt{Test} subsets based on the specified \texttt{Test Data Size}, preserving the time order of samples.

\textbf{Step 2: Sequence Creation} \\
\For{\textbf{each} data subset (\texttt{Train} and \texttt{Test})}{
    \For{\textbf{each} time step \(i\) in data subset}{
        Create a sequence from time step \(i\) to \(i + \texttt{Sequence Length} - 1\) \\
        Assign the target value for each sequence as the value at \(i + \texttt{Sequence Length}\) to align with next-step target
    }
}

\textbf{Step 3: Sequence Reshaping for \lstm} \\
Reshape both \texttt{Train} and \texttt{Test} sequences to match the \lstm~input format:

(\texttt{Batch size}, \texttt{Sequence Length}, \texttt{number of features})

\textbf{Step 4: Data Loader Initialization} \\
\For{\textbf{each} reshaped data subset (\texttt{Train} and \texttt{Test})}{
    Create a data loader with the specified \texttt{Batch Size} \\
    Set \texttt{shuffle} to \texttt{True} for training and \texttt{False} for testing
}

\KwRet{Training sequences, Testing sequences, and their corresponding data loaders}
\end{algorithm}


\subsection{Hardware setup}

The training of our deep learning models was conducted on high-performance hardware to handle the computational demands of wind power simulation. Specifically, we utilized a single \texttt{NVIDIA A100 GPU}, equipped with $80$ \texttt{GB} of Video Random Access Memory (\texttt{VRAM}) \footnote{\texttt{VRAM} refers to the dedicated memory on a \texttt{GPU} used to store data that it needs to process and render tasks efficiently. Unlike standard \texttt{RAM}, which is used by the \texttt{CPU}, \texttt{VRAM} is optimized for handling high-speed operations related to graphics and parallel computations, such as deep learning, simulations, and large matrix operations.}, optimized for intensive machine learning tasks. This configuration included a memory allocation of $256$ \texttt{GB} per \texttt{CPU} core to support large-scale data processing. The training job is managed through Simple Linux Utility for Resource Management (\texttt{SLURM}), using job-specific constraints to ensure compatibility with the \texttt{A100 GPU}'s extensive memory capacity, essential for efficient parallel processing and model training stability\footnote{\texttt{SLURM} is a widely used open-source workload manager and job scheduling system for high-performance computing (\texttt{HPC}) environments. Originally developed by the Lawrence Livermore National Laboratory, \texttt{SLURM} manages and allocates resources such as \texttt{CPU}s, \texttt{GPU}s, and memory across computing nodes in an \texttt{HPC} cluster, allowing users to submit and schedule jobs efficiently.}. This advanced hardware setup enabled effective training of our models, enhancing both speed and accuracy in handling the complex temporal dependencies within the wind power data.

\subsection{Code and data availability statement.}
The \texttt{CADNN} model is implemented using the \texttt{PyTorch} framework, taking advantage of its advanced tools for constructing and training deep learning models. \texttt{CADNN} is openly available to enable reproducibility and facilitate further research in wind power simulation, leveraging climate datasets and \dnn s for accurate, climate-informed predictions at this github repository\footnote{\url{https://github.com/Ali-Forootani/neural_wind_model}}. The code for data extraction, interpolation, statistical analysis, visualization , \dnn~training and evaluation is provided in the associated repository. It includes detailed comments and instructions for reproducing the results.
Additionally, a version will be archived on Zenodo\footnote{\url{https://zenodo.org/account/settings/github/repository/Ali-Forootani/neural_wind_model}, DOI: 10.5281/zenodo.14281375} for reference.

\subsection{\dnn~simulation setups}

In this study, we explore three deep learning architectures to forecast wind power generation, comparing their abilities to model complex temporal patterns in time-series data. The first scenario employs a structured \mlp~model based on the Sinusoidal Representation Network (\texttt{SIREN}) architecture, designed for high-frequency signal representation \citet{sitzmann2020implicit, forootani2024gn, forootani2023robust}. This model leverages sinusoidal weight initialization to capture fine-grained temporal dependencies. The second scenario implements a \lstm~network, a recurrent architecture that excels at learning sequential data patterns over long periods. In the third scenario, the deep learning model integrates both \lstm~and \trans~layers, allowing it to leverage temporal dependencies and self-attention for modeling long-range dependencies. While all models share the same input and output configurations—an input of $5$ features and a single-output prediction—each architecture is uniquely tailored to address the challenges of time-series forecasting. By comparing the \texttt{SIREN}-based \mlp, \lstm, and \trans-enhanced \lstm~approaches, we aim to identify the most effective structure for capturing wind power patterns, optimizing both forecasting accuracy and training stability. Algorithm \cref{alg:training_loop_dnn_model} summarizes the main training loop of the \dnn~structure considered in this article. In addition, out of total number of $679296$ samples (as mentioned earlier $2928 \times 232$), $90\%$ is randomly chosen for training the \dnn~structures and remaining $10\%$ for testing the results.

\paragraph{\mlp~based \texttt{CADNN}}
In the first scenario, we implement an \mlp~type \dnn~model using a structured, layered architecture tailored for wind power simulation. Our model configuration includes an input size of $5$ features and a single-output prediction, passing through a sequence of six hidden layers, each with $128$ hidden units. The model’s core is built around the \texttt{SIREN} (Sinusoidal Representation Network) architecture \citet{sitzmann2020implicit, forootani2023robust, forootani2024gn, forootani2024gs}, known for its ability to represent high-frequency signals, making it particularly effective for time-series data in wind forecasting. The network architecture initializes the model parameters with a specific sinusoidal weight initialization method, which allows the model to capture fine-grained temporal patterns. Number of epochs is considered $30000$ with learning rate $1e^{-5}$. Optimization is performed using the Adam optimizer with a weight decay of \(1 \times 10^{-6}\) to prevent overfitting. Additionally, we employ a Cyclic Learning Rate (\texttt{CLR}) scheduler in exponential range mode, which oscillates the learning rate between a base and maximum value, enhancing convergence by adjusting the learning rate dynamically. This architecture, optimized for periodic data, allows the model to leverage temporal dependencies effectively, providing a robust foundation for forecasting applications in renewable energy systems.

\begin{figure}[h]
    \centering
    \includegraphics[width=1\linewidth]{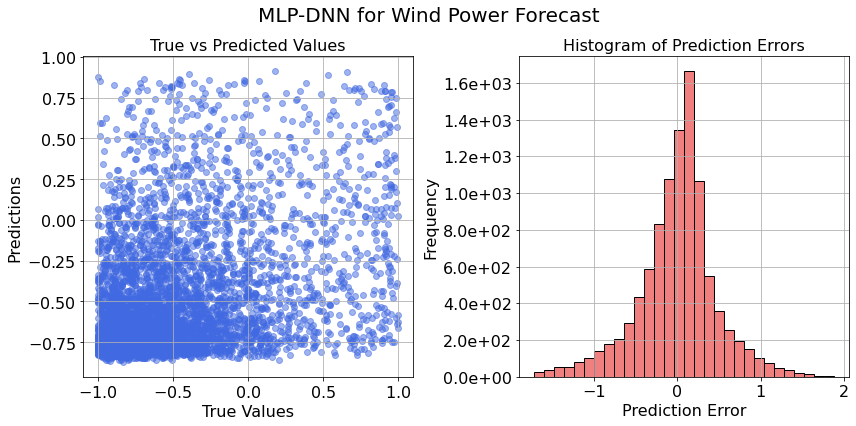}
    \caption{Prediction errors versus ground truth values in \mlp-\dnn.}
    \label{histogram_mlp}
\end{figure}

The scatter plot in \cref{histogram_mlp} illustrates the correlation between the true and predicted values generated by the \mlp-\dnn~model for wind power simulation. Points closely aligned along the diagonal line indicate strong prediction accuracy, as the predicted values closely match the true values. It
reveals a significant disparity between the predicted and true values, indicating that the \mlp-\dnn~struggles to accurately forecast wind power. Many predictions deviate considerably from the expected values, suggesting potential shortcomings in the model's ability to capture the underlying patterns in the data.

The histogram of prediction errors in \cref{histogram_mlp} presents the distribution of prediction discrepancies, defined as the difference between true and predicted values. A narrow centered distribution around zero with minimal spread indicates accurate predictions. The histogram of prediction errors further illustrates the model's limitations, as it shows a wide spread of errors rather than clustering around zero. This indicates that the \mlp-\dnn~frequently makes large errors in its predictions, highlighting the need for model refinement or alternative approaches to improve forecasting accuracy.

\begin{figure}[h]
    \centering
    \includegraphics[width=0.65\linewidth]{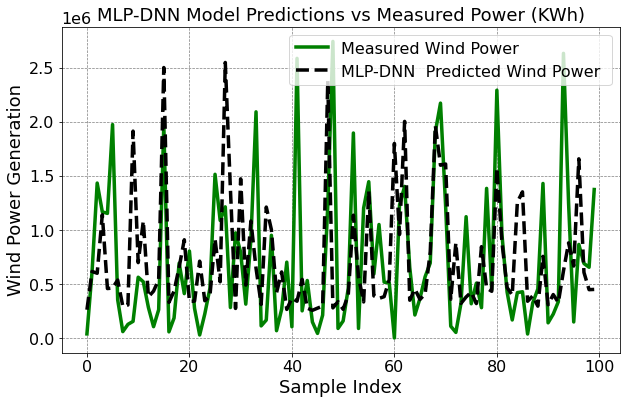}
    \caption{Comparing \mlp~based \dnn~versus true measurements for wind power simulation. It is worth to highlight that the true measurement model came came from \citet{lehneis2023temporally}.}
    \label{mlp_vs_tru}
\end{figure}

The plot \cref{mlp_vs_tru} illustrates the discrepancy between measured wind power generation and the \mlp-\dnn~model's predictions over the selected sample interval, indicating that the model struggles to accurately capture the variations in wind power output over the selected sample interval.

\paragraph{\lstm~based \texttt{CADNN}}
In the second scenario the deep learning architecture employed in our study consists of a \lstm~network tailored for time-series forecasting. Specifically, the model is constructed with an input size of $5$ features, passing sequentially through $6$ stacked \lstm~layers, each containing $128$ hidden units. Each \lstm~layer is initialized using Xavier initialization to optimize weight distribution. Following the \lstm~layers, a fully connected layer maps the output of the final \lstm~layer to a single prediction, as defined by the output size of one. This architecture captures complex temporal dependencies in the data across multiple layers, enhancing its ability to model long-range patterns. We use the Adam optimizer with a learning rate scheduler that dynamically reduces the learning rate based on validation loss, ensuring stability in training and adaptability to diminishing gradients over time. Number of epochs is considered $30000$ with initial learning rate $1e^{-3}$. This \lstm-based deep model is well-suited for our wind power simulation task due to its capacity to learn sequential dependencies and generate accurate time-step predictions.

\begin{figure}[h]
    \centering
    \includegraphics[width=1\linewidth]{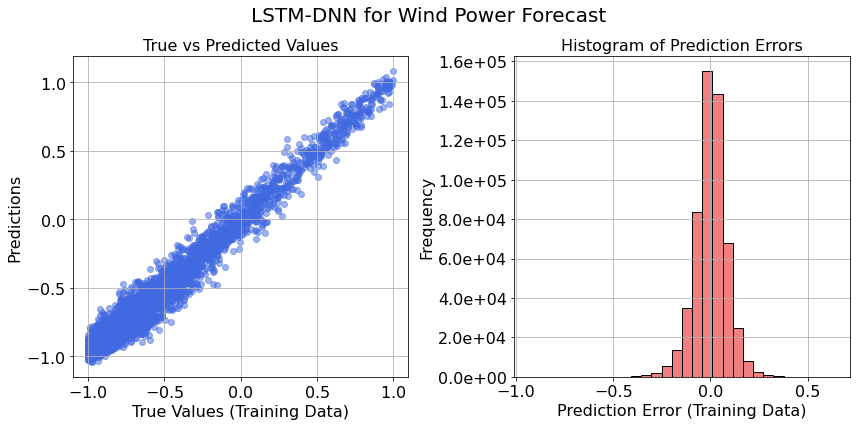}
    \caption{Prediction errors versus ground truth values in \lstm-\dnn.}
    \label{hist_lstm}
\end{figure}

The scatter plot in \cref{hist_lstm} demonstrates a strong correlation between the true values and the predicted values, with most predictions closely aligning along the diagonal line, indicating that the \lstm-\dnn~model performs exceptionally well in accurately forecasting outcomes.

The histogram of prediction in \cref{hist_lstm} errors reveals a tight distribution centered around zero, suggesting that the \lstm-\dnn~model consistently produces accurate predictions, with only a few instances of significant deviation, further highlighting its effectiveness in the task at hand.

\begin{figure}[h]
    \centering
    \includegraphics[width=0.65\linewidth]{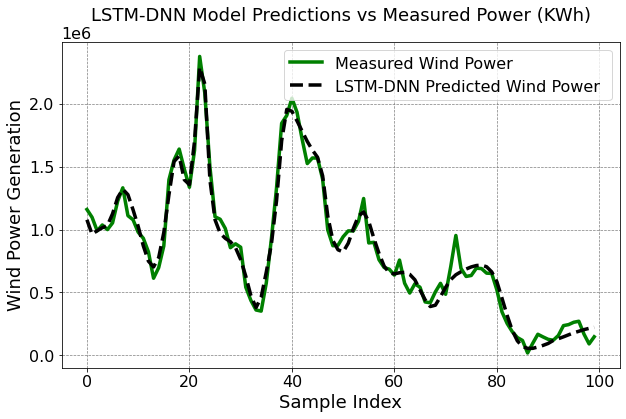}
    \caption{Comparing \texttt{LSTM} based \texttt{DNN} versus true measurements for wind power power simulation. It is worth to highlight that the true measurement model came came from \citet{lehneis2023temporally}.}
    \label{lstm_vs_true}
\end{figure}

\begin{figure}[h]
    \centering
    \includegraphics[width=0.65\linewidth]{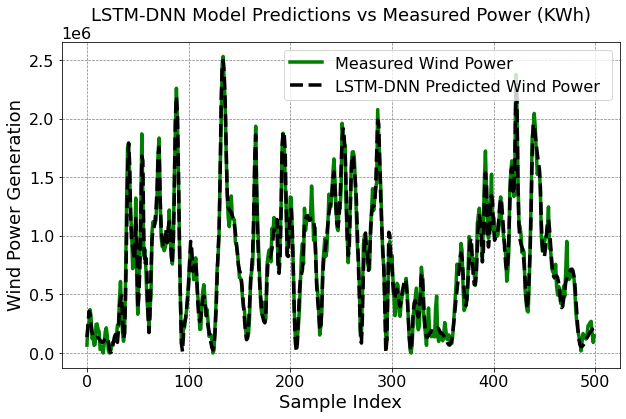}
    \caption{Comparing \texttt{LSTM} based \texttt{DNN} versus true measurements for wind power simulation. It is worth to highlight that the true measurement model came came from \citet{lehneis2023temporally}.}
    \label{lstm_vs_true_2}
\end{figure}

The line plots in \cref{lstm_vs_true} and \cref{lstm_vs_true_2} vividly illustrate the performance of the \lstm-\dnn~model in predicting wind power generation, with the predicted values closely following the trend of the measured wind power, especially within the selected sample range. This indicates a strong predictive capability of the model, as evidenced by the minimal divergence between the predicted and true values.

\paragraph{\lstm-\trans~based \texttt{CADNN}}
In this scenario, the deep learning architecture employed for time-series forecasting integrates both \lstm~and \trans~\\ layers. Specifically, the model is designed with an input size of $5$ features, and it passes through $6$ stacked \lstm~layers, each containing $64$ hidden units. The \lstm~ layers are initialized using Xavier initialization to ensure effective weight distribution. Following the \lstm~ layers, the output is passed through $2$ \trans~ layers, each utilizing $4$ attention heads and a feed-forward network of size $20$. This hybrid architecture enables the model to leverage both the temporal dependencies captured by the \lstm~layers and the self-attention mechanism provided by the \trans~ layers, improving its ability to model complex sequential patterns with long-range dependencies. The final output is generated by a fully connected layer that maps the output of the last \trans~ layer to a single prediction, as defined by the output size of one. The Adam optimizer is used with an initial learning rate of $1e^{-3}$, and a learning rate scheduler is applied to reduce the learning rate dynamically based on validation loss, ensuring training stability and adaptability. The model is trained for $30000$ epochs, making it particularly well-suited for forecasting tasks that involve both short-term and long-term temporal dependencies, such as wind power simulation.

\begin{figure}[htp]
    \centering
    \includegraphics[width=1\linewidth]{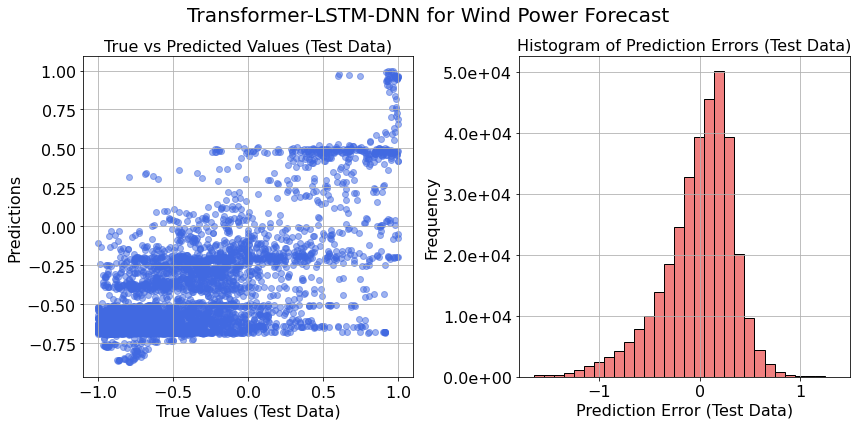}
    \caption{Prediction errors versus ground truth values in \trans-based \lstm-\dnn.}
    \label{hist_transform_lstm}
\end{figure}

The scatter plot in \cref{hist_transform_lstm} illustrates a weak correlation between the true values and predicted values, with the majority of predictions clustering not closely along the diagonal line. This alignment indicates that the \lstm-\trans-\dnn~model achieves low accuracy in forecasting wind power outputs.

Additionally, the histogram of prediction errors in \cref{hist_transform_lstm} shows a less concentrated distribution around zero, implying that the model consistently can not provides precise predictions, with maximal occurrences of substantial error. This performance underscores the effectiveness of the \lstm-\trans-\dnn~architecture for accurate time-series forecasting in wind power applications.

\begin{figure}[htp]
    \centering
    \includegraphics[width=0.65\linewidth]{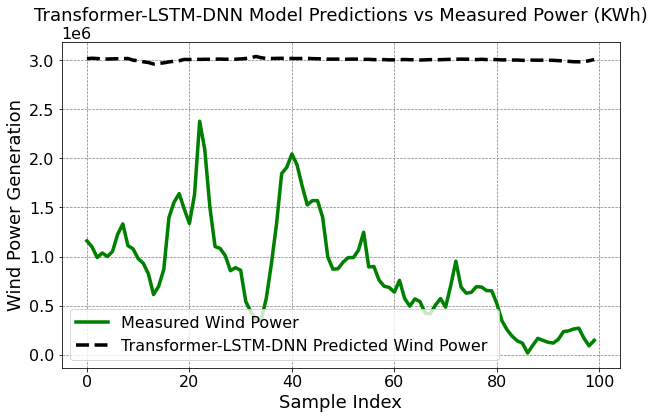}
    \caption{Comparing \trans-\texttt{LSTM} based \texttt{DNN} versus true measurements for wind power production forecast}
    \label{transform_lstm_vs_true}
\end{figure}

\begin{figure}[htp]
    \centering
    \includegraphics[width=0.65\linewidth]{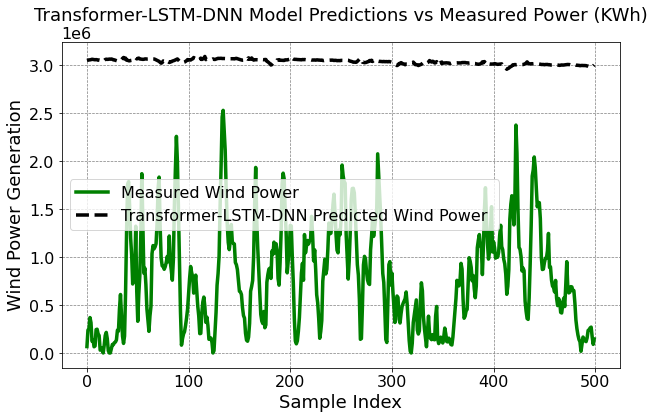}
    \caption{Comparing \trans-\texttt{LSTM} based \texttt{DNN} versus true measurements for wind power simulation. It is worth to highlight that the true measurement model came came from \citet{lehneis2023temporally}.}
    \label{transform_lstm_vs_true_2}
\end{figure}

\begin{figure}[htp]
    \centering
    \includegraphics[width=0.65\linewidth]{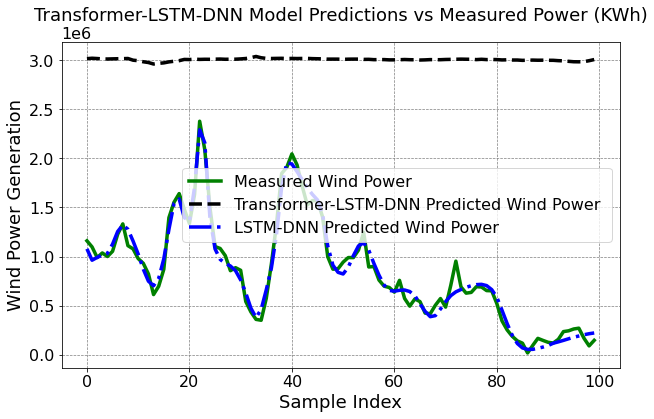}
    \caption{Comparing \trans-\texttt{LSTM} based \dnn~versus true measurements and \lstm based \dnn~for wind power simulation. It is worth to highlight that the true measurement model came came from \citet{lehneis2023temporally}.}
    \label{transform_lstm_vs_true_vs_lstm}
\end{figure}

The plots in \cref{transform_lstm_vs_true}, \cref{transform_lstm_vs_true_2}, and \cref{transform_lstm_vs_true_vs_lstm} highlight discrepancies between the measured wind power generation and the \lstm-\trans-\dnn~model's predictions over the selected sample interval. These inconsistencies suggest that the model struggles to capture the full extent of variability in wind power output.


\begin{algorithm}
\caption{General Training Loop for Deep Learning Model}
\label{alg:training_loop_dnn_model}
\KwIn{\texttt{DNN} Model $\mathcal{M}$, Optimizer $\mathcal{O}$ (e.g., Adam optimizer), Learning rate scheduler \texttt{scheduler}, Loss function $\mathcal{L}$, Number of epochs $E$, Training dataset loader $\mathcal{D}_{train}$}
\KwOut{Trained model parameters $\theta^*$, total loss over epochs $\mathcal{L}_{total}$}

Initialize empty list $\mathcal{L}_{total} \gets []$ \tcp{To store total losses}

\For{epoch $e \gets 1$ \KwTo $E$} {
    $loss_{data} \gets 0$ \;
    $t_{start} \gets \texttt{time.time()}$ \tcp{Record start time}
    
    $loss_{data} \gets \mathcal{L}(\mathcal{D}_{train}, \mathcal{M})$ \tcp{Evaluate loss on train data using the model}
    $\mathcal{L} \gets loss_{data}$ \tcp{Compute the total loss}

    $\mathcal{L}_{total}.\texttt{append}(\mathcal{L})$ \tcp{Store total loss}

    \textbf{Gradient Calculation:} \;
    $\mathcal{O}.\texttt{zero\_grad()}$ \tcp{Zero the gradients}
    $\mathcal{L}.\texttt{backward()}$ \tcp{Backpropagation step}
    $\mathcal{O}.\texttt{step()}$ \tcp{Update model parameters}

    \textbf{Learning Rate Scheduler:} \;
    \texttt{scheduler.step()} \tcp{Adjust the learning rate}
    
    \textbf{Logging (Optional):} \;
    Output current progress: epoch $e$ and loss $\mathcal{L}$
}
\KwRet $\theta^*, \mathcal{L}_{total}$ \tcp{Return trained model parameters and total loss over epochs}
\end{algorithm}


\section{Conclusion}\label{conclude}
This study made several important contributions to the literature on renewable energy forecasting by integrating large-scale CMIP6 climate data with advanced machine learning models for wind power prediction. We proposed a novel data processing pipeline that included temporal resampling, spatial interpolation, and normalization techniques, establishing best practices for adapting climate datasets for machine learning frameworks. Our comparative analysis of deep learning models demonstrated that \lstm~networks were highly effective for this application, significantly outperforming \mlp s and \trans-enhanced \lstm~models in capturing the temporal dependencies essential for accurate forecasts.

In addition, we provided a dedicated \texttt{Python} package built with \texttt{PyTorch} to support the reproducibility of our framework. This package offered modules for data preparation, model training, and evaluation, making it a practical resource for researchers and practitioners aiming to implement climate-based renewable energy forecasting models. By making this tool available, we fostered further research and application in this domain, supporting scalable and efficient forecasting methods.

This work advanced the field by providing a rigorous, adaptable framework that combined climate science with machine learning, addressing a critical need in renewable energy forecasting. The methodology and tools presented here not only enhanced wind power prediction accuracy but also laid the foundation for future innovations in climate data utilization for sustainable energy applications.

\section*{Acknowledgment}
This study is funded by the \href{https://man0euvre.eu/}{\emph{Man0EUvRE}} (100695543), which is co-financed by means of taxation based on the budget adopted by the representatives of the Landtag of Saxony. ``Man0EUvRE – Energy System Modelling for Transition to a net-Zero 2050 for EU via REPowerEU,'' is funded by CETPartnership, the European Partnership under Joint Call 2022 for research proposals, co-funded by the European Commission (GA N°101069750).


\bibliographystyle{elsarticle-num-names} 
\bibliography{reference}






\end{document}